\documentclass[11pt]{article}

\usepackage[final]{acl}

\usepackage{times}
\usepackage{latexsym}
\usepackage{multirow}
\usepackage{booktabs}
\usepackage{subcaption} 
\usepackage{listings}
\usepackage{longtable}
\usepackage{tcolorbox}
\usepackage{colortbl}

\usepackage[T1]{fontenc}

\usepackage{arabtex}
\usepackage{utf8}
\setcode{utf8}
\usepackage{microtype}

\usepackage{inconsolata}

\usepackage{graphicx}
\usepackage{enumitem} 
\title{\textit{HalluVerse25}: Fine-grained Multilingual Benchmark Dataset for LLM Hallucinations}

\author{Samir Abdaljalil \\
  Texas A\&M University\\
  College Station, TX., USA\\
  \texttt{sabdaljalil@tamu.edu} \\ \And
  Hasan Kurban \\
  Hamad Bin Khalifa University \\
  Doha, Qatar\\ 
  \texttt{hkurban@hbku.edu.qa} \\ \And
  Erchin Serpedin \\
  Texas A\&M University\\
  College Station, TX., USA\\
  \texttt{eserpedin@tamu.edu} \\}

\begin{document}

\maketitle

\begin{center}
    {\large \textbf{Abstract}} 
\end{center}

\begingroup
\setlength{\leftskip}{0.5cm}  
\setlength{\rightskip}{0.5cm} 
\noindent Large Language Models (LLMs) are increasingly used in various contexts, yet remain prone to generating non-factual content, commonly referred to as “hallucinations''. The literature categorizes hallucinations into several types, including entity-level, relation-level, and sentence-level hallucinations.  
However, existing hallucination datasets often fail to capture fine-grained hallucinations in multilingual settings. In this work, we introduce \textit{HalluVerse25}, a multilingual LLM hallucination dataset that categorizes fine-grained hallucinations in English, Arabic, and Turkish. Our dataset construction pipeline uses an LLM to inject hallucinations into factual biographical sentences, followed by a rigorous human annotation process to ensure data quality.  
We evaluate several LLMs on \textit{HalluVerse25}, providing valuable insights into how proprietary models perform in detecting LLM-generated hallucinations across different contexts.
\par
\endgroup

\section{Introduction}
Large Language Models (LLMs) have demonstrated promising results in various natural language processing (NLP) tasks, including summarization \cite{liu-etal-2024-learning, ramprasad-etal-2024-analyzing}, machine translation \cite{zhu-etal-2024-multilingual}, question answering \cite{kamalloo-etal-2023-evaluating}, and many more. Despite this, the phenomenon of producing fabricated and nonfactual content, known as "hallucinations'', has also been on the rise \cite{banerjee2024llmshallucinateneedlive}. Hallucinations can lead to the spread of misinformation, as they can sometimes appear to be plausible, particularly to individuals who are unfamiliar with the subject matter being generated \cite{berberette2024redefininghallucinationllmspsychologyinformed}. As a result, analyzing and detecting hallucinations has become a major focus in the literature. 


Several datasets have been developed to understand and analyze hallucinations generated by LLMs \cite{mubarak-etal-2024-halwasa, li-etal-2023-halueval, manakul-etal-2023-selfcheckgpt, mishra2024finegrained}. Existing datasets, however, mostly consist of monolingual data, primarily in English, and often present hallucination in a binary classification setting. Such datasets lack the fine-grained annotation necessary for a deeper understanding of hallucination types. To fill this gap, we introduce \textit{\textit{HalluVerse25}}, a fine-grained multilingual LLM hallucination dataset. 

Our main contributions are as follows:
\begin{itemize}
\item We introduce the first multilingual fine-grained LLM hallucination dataset. The dataset spans across three languages, namely English, Arabic, and Turkish, with 1310, 828, and 978 data samples, respectively. This dataset is made publicly available to advance research in the field\footnote{Available Upon Request}. 
\item We detail a comprehensive and reproducible dataset creation and annotation protocol.
\item We perform a detailed analysis of the performance of proprietary LLMs on detecting fine-grained hallucinations in multiple settings.
\end{itemize}

The remainder of the paper is organized as follows: Section \ref{sec:related} discusses relevant literature on hallucination datasets and benchmarks. Section \ref{sec:method} details the dataset construction methodology. Section \ref{sec:analysis} presents a comprehensive analysis of the dataset. Section \ref{sec:exp} discusses experimental results of several LLMs on the dataset. Finally, Section \ref{sec:conclusion} concludes the paper and outlines future directions. 

\section{Related Work}
\label{sec:related}
\paragraph{Hallucination in LLMs.} Hallucination in LLMs continues to be a pertinent issue with its diverse real-world applications \cite{li-etal-2024-dawn}. An increasing proportion of the literature explores this phenomenon \cite{10.1145/3703155, 10.1145/3571730}. 
A portion of the literature focuses on the underlying causes of hallucination \cite{mckenna-etal-2023-sources, das-etal-2022-diving}, while other works propose several methodologies to detect and mitigate LLM hallucinations \cite{chen2024inside, chen-etal-2024-unified-hallucination, varshney2023stitchtimesavesnine}. For instance, \citet{chen2024inside} use the internal hidden states of the LLM to detect the presence of hallucinations using an Eigenscore metric. While significant progress has been made in understanding hallucination in LLMs, the development of comprehensive benchmarks and datasets remains crucial for  evaluating and mitigating these phenomena.

\paragraph{Hallucination Benchmarks \& Datasets.}  To address the need for standardized and comprehensive evaluation, several benchmarks and datasets have been introduced in the literature, 
with the aim of capturing and quantifying hallucinations in various contexts. With 35,000 samples, HaluEval encompasses multiple tasks such as question answering, summarization, and open-ended generations in English \cite{li-etal-2023-halueval}, while \citet{mubarak-etal-2024-halwasa} introduce Halwasa, a dataset comprised of 10,000 sentence-level hallucination samples in Arabic, extending the scope to non-English languages. Other smaller datasets, such as the Wikibio hallucination dataset \cite{manakul-etal-2023-selfcheckgpt}, were introduced in the literature. In addition to binary hallucination datasets, fine-grained hallucination types have also been defined and explored. \citet{mishra2024finegrained} define several distinct types of factual errors including:
\begin{itemize}
\item Entity-level: Errors where an incorrect entity in a statement can be replaced to make it accurate; 
\item Relation: Errors caused by incorrect semantic relationships, such as verbs or prepositions;
\item Sentence-level: Errors where a large portion of the statement contradicts evidence.
\end{itemize}
Using these hallucination types, the authors introduced FAVABENCH, a dataset consisting of 1,000 hallucinated query responses with fine-grained labels. Other fine-grained non-English hallucination datasets have also been recently presented in the literature, such as ANAH \cite{ji-etal-2024-anah}, a bilingual Chinese/English sentence-level annotated hallucination dataset that covers a variety of topics. 

Although progress has been made, a gap still exists in the literature for a self-contained, fine-grained multilingual hallucination dataset. Existing benchmarks mainly focus on individual languages, often limited to English, or lack the fine-grained annotations needed to explore and understand the nuances of hallucination across different languages. To fill this gap, we present \textit{HalluVerse25}, a fine-grained multilingual LLM hallucination dataset, which aims to provide in-depth insight into different types of hallucinations in multilingual contexts.

\section{Constructing \textit{HalluVerse25}}
\label{sec:method}

\begin{figure*}[ht]
\centering
\includegraphics[width=\textwidth]{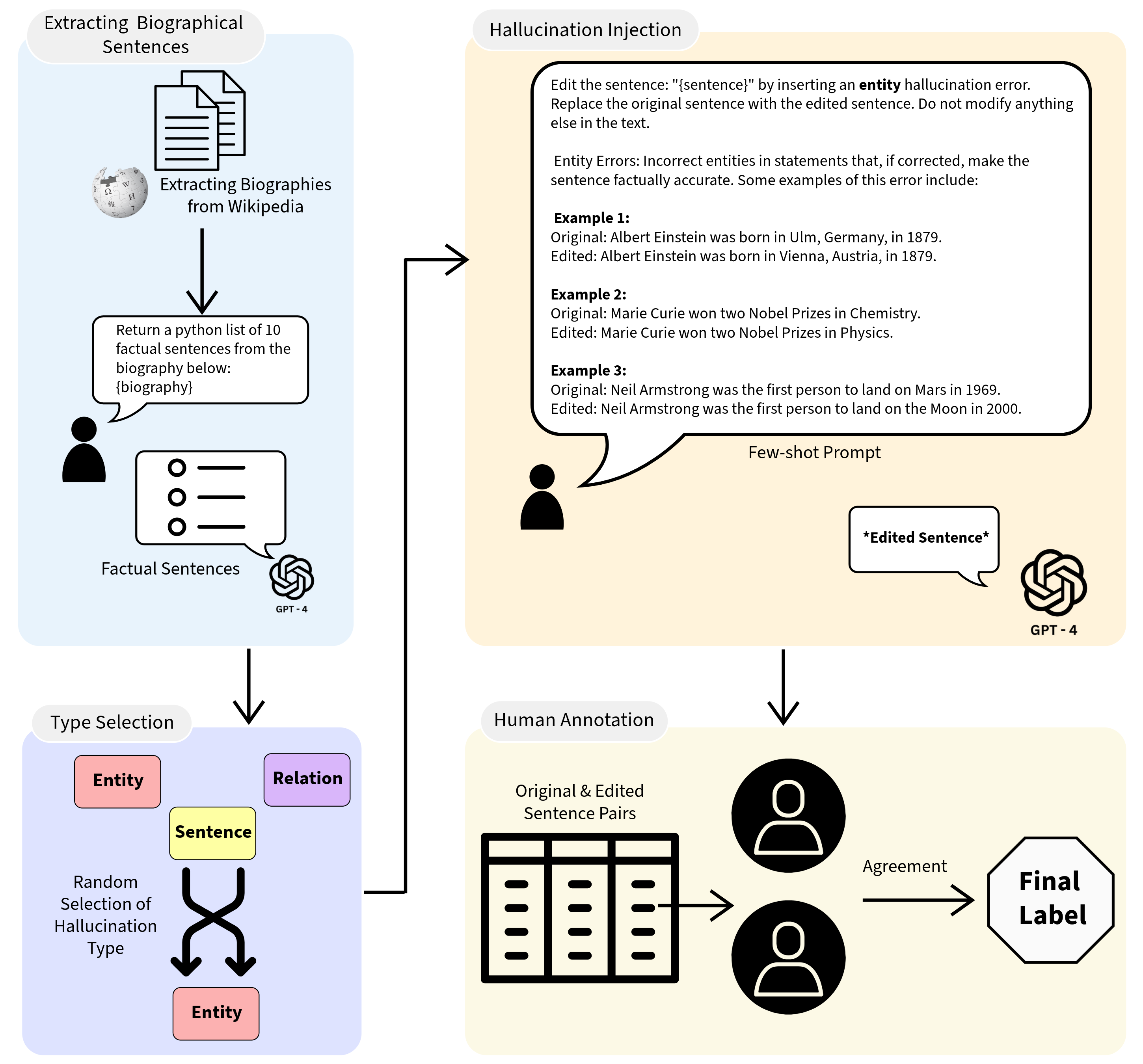} 
\caption{Pipeline of Dataset Construction, including extraction of biographical factual sentences, automatic injection of fine-grained hallucinations, and human annotation.}
\label{fig:method}
\end{figure*}

The main objective behind the construction of this dataset is to understand what different types of hallucinations look like in multiple languages. The dataset includes the original factual sentences and their hallucinated versions, as well as the hallucination type associated with each edited sentence. The construction pipeline is illustrated in Fig. \ref{fig:method}.
Furthermore, samples from the final dataset are shown in Table \ref{tab:samples}.

\subsection{Extraction of Biographical Data}

\paragraph{Wikipedia Links from Wikidata.} 
For this work, we focus primarily on building a data set based on biographical data. The required data is extracted from Wikidata using a SPARQL query. Wikidata is a multilingual knowledge graph by the Wikimedia Foundation \cite{abdaljalil-mubarak-2024-wikidata}. The query identifies 500 human entities from the Wikidata knowledge base (specifically those represented by the entity identifier \texttt{wd:Q5}). For each of these entities, the query retrieves the associated Wikipedia links in three languages: English, Arabic, and Turkish. The query uses Wikidata Property \texttt{wdt: P2025} to obtain the respective links to the articles in these languages. The final output of the query provides a list of entities with their corresponding Wikipedia links in the specified languages, which are then used for further data extraction. 
The full SPARQL query used to extract this data is provided in Appendix \ref{app:sparql}.

\paragraph{Extracting Factual Sentences.} 
The next step involved accessing the links and extracting the text from each page using BeautifulSoup\footnote{https://pypi.org/project/beautifulsoup4/}. First, we excluded pages that contained fewer than 20 sentences of text. This included pages that had minimal textual content, or did not contain any. This resulted in the following number of pieces of text from the 500 original entities for each language: 324, 214, and 260, for English, Arabic, and Turkish, respectively.

Then, GPT-4 was instructed to extract five factual sentences from each of the valid texts. This step was important to inject hallucinations into the text, ensuring that we only kept sentences containing factual information. By doing so, we were able to disregard any sentences that did not include verifiable factual content. The instruction for extracting factual sentences can be found in Table \ref{tab:instruct_fact} in Appendix \ref{app:prompts}.

\subsection{Automated Injection of Hallucination}
Since we are exploring different types of hallucinations - namely entity, relation, and sentence - we used a random generator to decide which type of hallucination would be injected into each sentence. This approach helped maintain diversity in the outputs included in the dataset, which ensured a balanced representation of hallucination types, preventing bias toward any particular category.

Once the type of hallucination was selected, we used a three-part few-shot prompting technique to instruct GPT-4 to inject the chosen hallucination type. The first part of the prompt specified the type of hallucination to be injected. The second part provided a clear definition of the type of hallucination selected, ensuring that GPT-4 understood the context and main aim. The third part included three pairs of original and hallucinated sentences, offering clear examples to guide the model's behavior. This structure helped maintain accuracy and consistency in the generated outputs. The prompts used for this process can be found in Tables~\ref{tab:entity_prompt}, \ref{tab:relation_prompt}, and \ref{tab:sentence_prompt} in Appendix \ref{app:prompts}. 

\renewcommand{\arraystretch}{1.5} 

\begin{table*}[htbp]
    \centering
    \caption{Samples from \textit{HalluVerse25} for English (En), Arabic (Ar), and Turkish (Tr), illustrating different types of hallucinations, including entity (Ent.), relation (Rel.), and sentence (Sent.) hallucinations. The original and edited samples demonstrate how these hallucinations alter the meaning of text. English translation for Arabic and Turkish data is provided.}
    \resizebox{\textwidth}{!}{ 
    \begin{tabular}{c c c c}
    \toprule
    \textbf{Lang.} & \multicolumn{2}{c}{\textbf{Sample}} & \textbf{Type} \\
    \midrule
    \multirow{6}{*}{\centering \textbf{En}} 
    & \textbf{Orig.}  & A long time leader within the Nahdlatul Ulama organization, he was the founder of the \textcolor{red}{National Awakening Party (PKB)}.  & \\
    & \textbf{Edited} & A long time leader within the Nahdlatul Ulama organization, he was the founder of the \textcolor{red}{Democratic Party of Japan (DPJ)}.  & Ent.  \\
    \cmidrule(lr){2-4}
    & \textbf{Orig.}  & After the war's tide decisively turned \textcolor{red}{against} Japan, Tojo resigned as prime minister on 18 July 1944.  & \\
    & \textbf{Edited} & After the war's tide decisively turned \textcolor{red}{with} Japan, Tojo resigned as prime minister on 18 July 1944.  & Rel.  \\
    \cmidrule(lr){2-4}
    & \textbf{Orig.}  & Abd Allah had a brother, Utba, and \textcolor{red}{at least two wives in} Muhammad's lifetime.  & \\
    & \textbf{Edited} & Abd Allah had a brother, Utba, and \textcolor{red}{won a Nobel Prize for Physics during} Muhammad's lifetime.  & Sent.  \\
    \midrule
    \midrule
    \multirow{6}{*}{\centering \textbf{Ar}} 
    & \textbf{Orig.}  & \begin{tabular}[c]{@{}c@{}}\<أرسل عبد المجيد الأول خمسة سفن إلى أيرلندا كجزء من عملية إغاثة عاجلة خلال المجاعة.> \\ (Abdulmejid I sent five ships to \textcolor{red}{Ireland} as part of an urgent relief operation during the famine.)\end{tabular} & \\
    & \textbf{Edited} & \begin{tabular}[c]{@{}c@{}}\<أرسل عبد المجيد الأول خمسة سفن إلى نيوزيلندا كجزء من عملية إغاثة عاجلة خلال المجاعة.> \\ (Abdulmejid I sent five ships to \textcolor{red}{New Zealand} as part of an urgent relief operation during the famine.)\end{tabular} & Ent.  \\
    \cmidrule(lr){2-4}
    & \textbf{Orig.}  & \begin{tabular}[c]{@{}c@{}}\<ألان بين  هو أحد رواد الفضاء الأمريكيين لذين هبطوا على القمر خلال رحلات برنامج أبولو.> \\ (Alan Bean is one of the American astronauts who landed on the Moon \textcolor{red}{during} the Apollo program missions.)\end{tabular} & \\
    & \textbf{Edited} & \begin{tabular}[c]{@{}c@{}}\<ألان بين  هو أحد رواد الفضاء الأمريكيين لذين هبطوا على القمر بعد رحلات برنامج أبولو.> \\ (Alan Bean is one of the American astronauts who landed on the Moon \textcolor{red}{after} the Apollo program missions.)\end{tabular} & Rel.  \\
    \cmidrule(lr){2-4}
    
    & \textbf{Orig.}  & \begin{tabular}[c]{@{}c@{}}\<استطاع الفرار من عقوبات حتمية بعد رجوع الحكم الملكي في عام 1660> \\ (He was able to \textcolor{red}{escape inevitable punishments} after the restoration of the monarchy in 1660)\end{tabular} & \\
    & \textbf{Edited} & \begin{tabular}[c]{@{}c@{}}\<استطاع الهبوط على سطح القمر بعد رجوع الحكم الملكي في عام 1660.> \\ (He was able to \textcolor{red}{land on the surface of the moon} after the restoration of the monarchy in 1660)\end{tabular} & Sent. \\
    \midrule
    \midrule
    \multirow{6}{*}{\centering \textbf{Tr}} & \textbf{Orig.}  & \begin{tabular}[c]{@{}c@{}}Kalabalık bir ailenin çocuğu olarak Rusya’nın Ryazan kentinde dünyaya geldi.\\ (He/She was born in the city of \textcolor{red}{Ryazan}, Russia, as a child of a large family.)\end{tabular} & \\
    & \textbf{Edited} & \begin{tabular}[c]{@{}c@{}}Kalabalık bir ailenin çocuğu olarak Rusya’nın Moskova kentinde dünyaya geldi.\\ (He/She was born in the city of \textcolor{red}{Moscow}, Russia, as a child of a large family.)\end{tabular} & Ent. \\
    \cmidrule(lr){2-4}
    & \textbf{Orig.}  & \begin{tabular}[c]{@{}c@{}}Bennington'un ebeveynleri, Bennington 11 yaşındayken ayrıldı.\\ (Bennington's parents \textcolor{red}{separated} when Bennington was 11 years old.)\end{tabular}  & \\
    & \textbf{Edited} & \begin{tabular}[c]{@{}c@{}}Bennington'un ebeveynleri, Bennington 11 yaşındayken bir araya geldi.\\ (Bennington's parents \textcolor{red}{got together} when Bennington was 11 years old.)\end{tabular}   & Rel.  \\
    \cmidrule(lr){2-4}
    & \textbf{Orig.}  & \begin{tabular}[c]{@{}c@{}}Bergman 2005 yılında Time dergisi tarafından dünyanın yaşayan en büyük yönetmeni olarak nitelendirilmiştir.\\ (In 2005, Bergman was described by Time magazine as \textcolor{red}{the greatest living director in the world.})\end{tabular}   & \\
    & \textbf{Edited} & \begin{tabular}[c]{@{}c@{}}Bergman 2005 yılında Time dergisi tarafından Mars'ta yaşam keşfeden ilk yönetmen olarak nitelendirilmiştir.\\ (In 2005, Bergman was described by Time magazine as \textcolor{red}{the first director to discover life on Mars.})\end{tabular}   & Sent.  \\
    \bottomrule
    \end{tabular}
    } 
    
    \label{tab:samples}
\end{table*}

\subsection{Human Annotation}
To validate the output data, human annotation was an important part of the process. For this, we instructed two native speakers, each with at least a bachelor's degree, to annotate the data for each language. The original and edited sentence pairs were presented to the annotators with no other information provided. We conducted a 45-minute training session to explain the annotation guidelines and shared an annotation protocol (found in Appendix \ref{app:protocol}) that detailed the types of hallucinations, along with examples corresponding to each type. Annotators were instructed to label a pair with 0 if there was grammatically incorrect content or if the edited sentence did not contain any hallucinations.

We report Cohen’s Kappa ($\kappa$) \cite{kappa} to
quantify the annotation agreement between the two human annotators for each language. We compute $\kappa$ on the full sample and obtain $\kappa$ of 0.748, 0.719, and 0.805, showing substantial agreement 
($0.60 \le  \kappa <  0.8$) for English and Arabic, and perfect agreement ($0.80 \le  \kappa \le  1.00$) for Turkish, respectively. To address disagreements, annotators discussed and collaboratively resolved discrepancies by reaching a consensus on a single label.


\section{Data Analysis}
\label{sec:analysis}

\begin{figure}[htbp]
    \centering
    
    \begin{subfigure}{\linewidth}
        \centering
        \includegraphics[width=\linewidth]{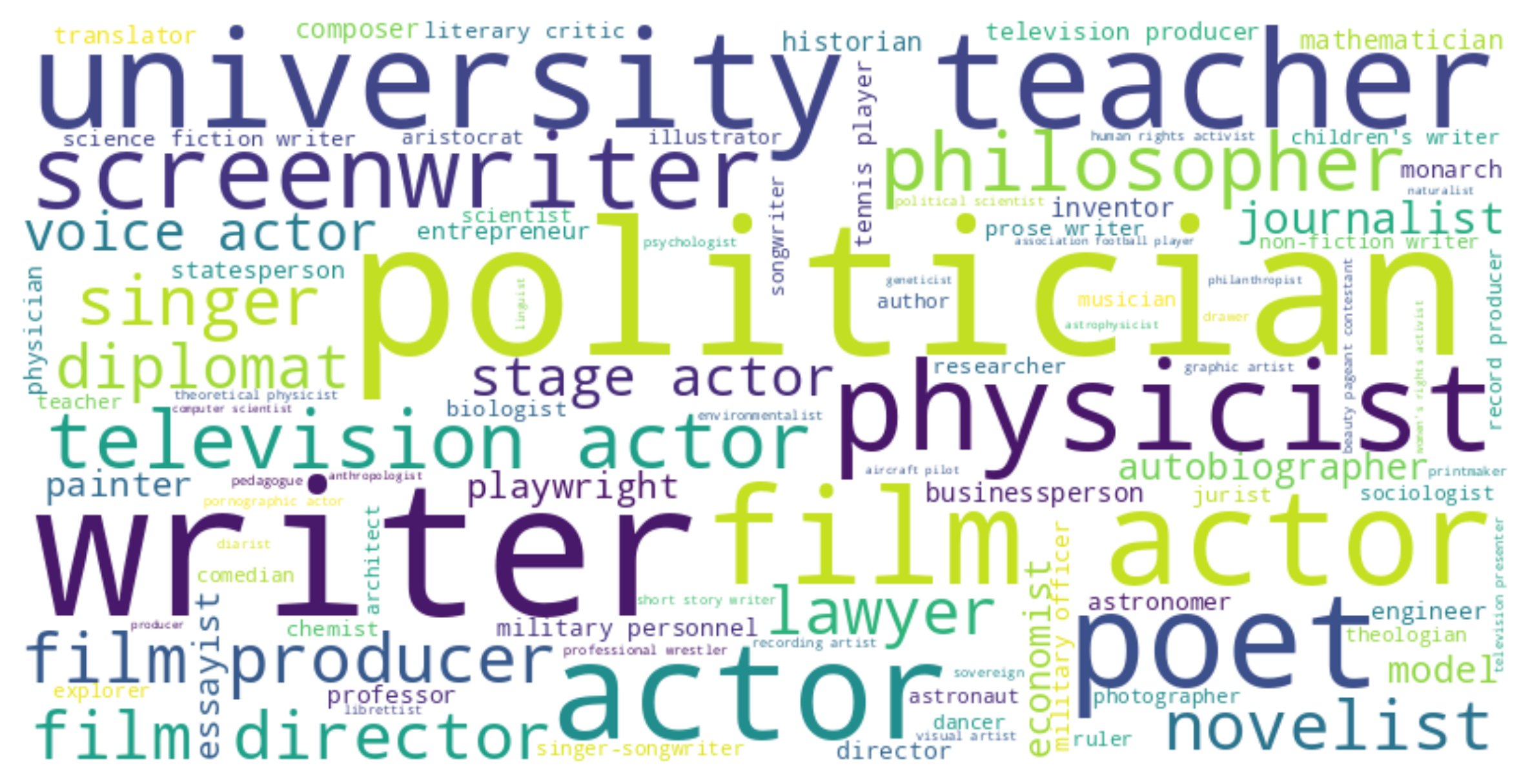}
        \caption{Profession}
        \label{fig:profession}
    \end{subfigure}
    
    \vspace{0.5cm} 
    
    \begin{subfigure}{\linewidth}
        \centering
        \includegraphics[width=\linewidth]{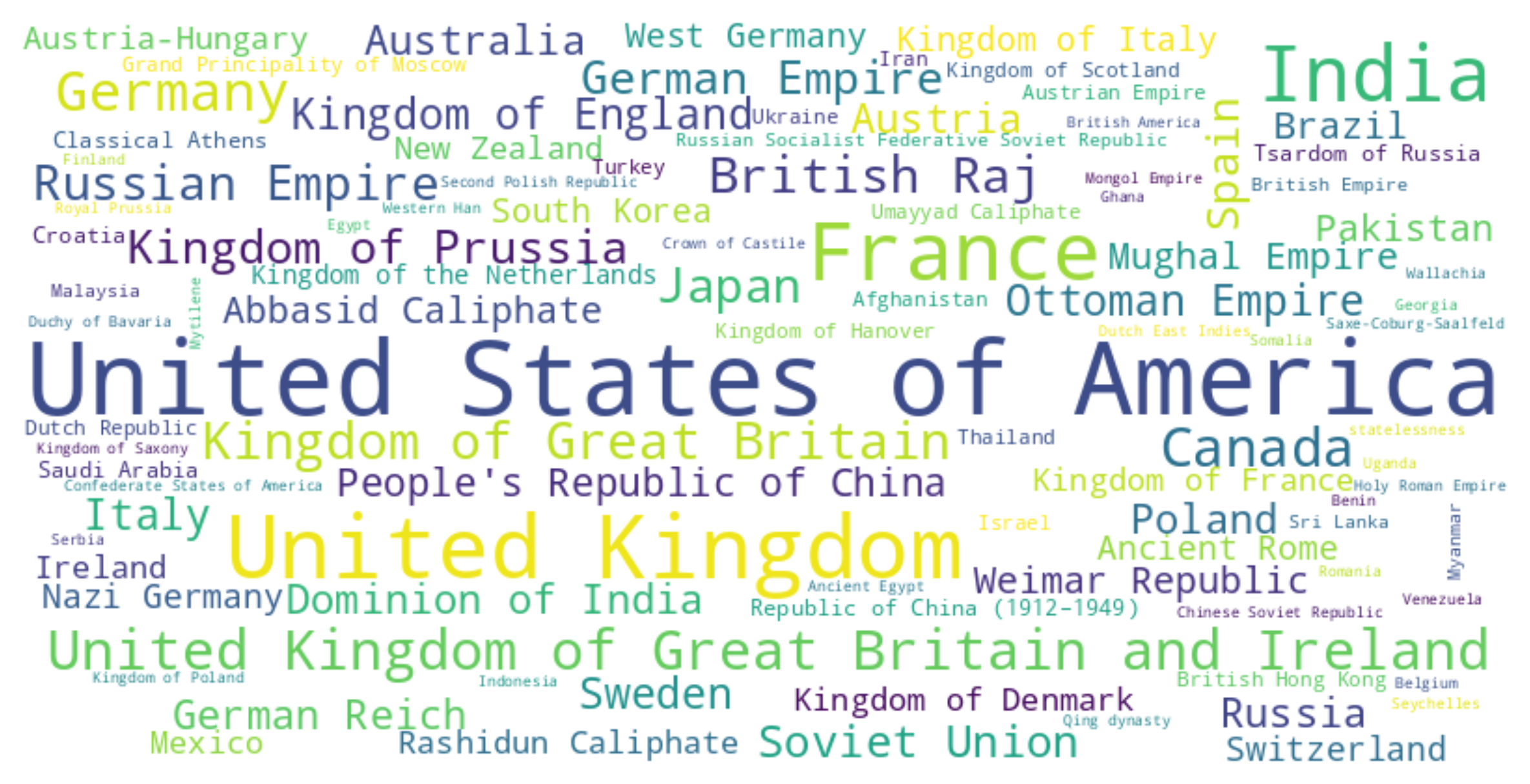}
        \caption{Country of Origin}
        \label{fig:country}
    \end{subfigure}
    
    \vspace{0.5cm} 
    
    \begin{subfigure}{\linewidth}
        \centering
        \includegraphics[width=\linewidth]{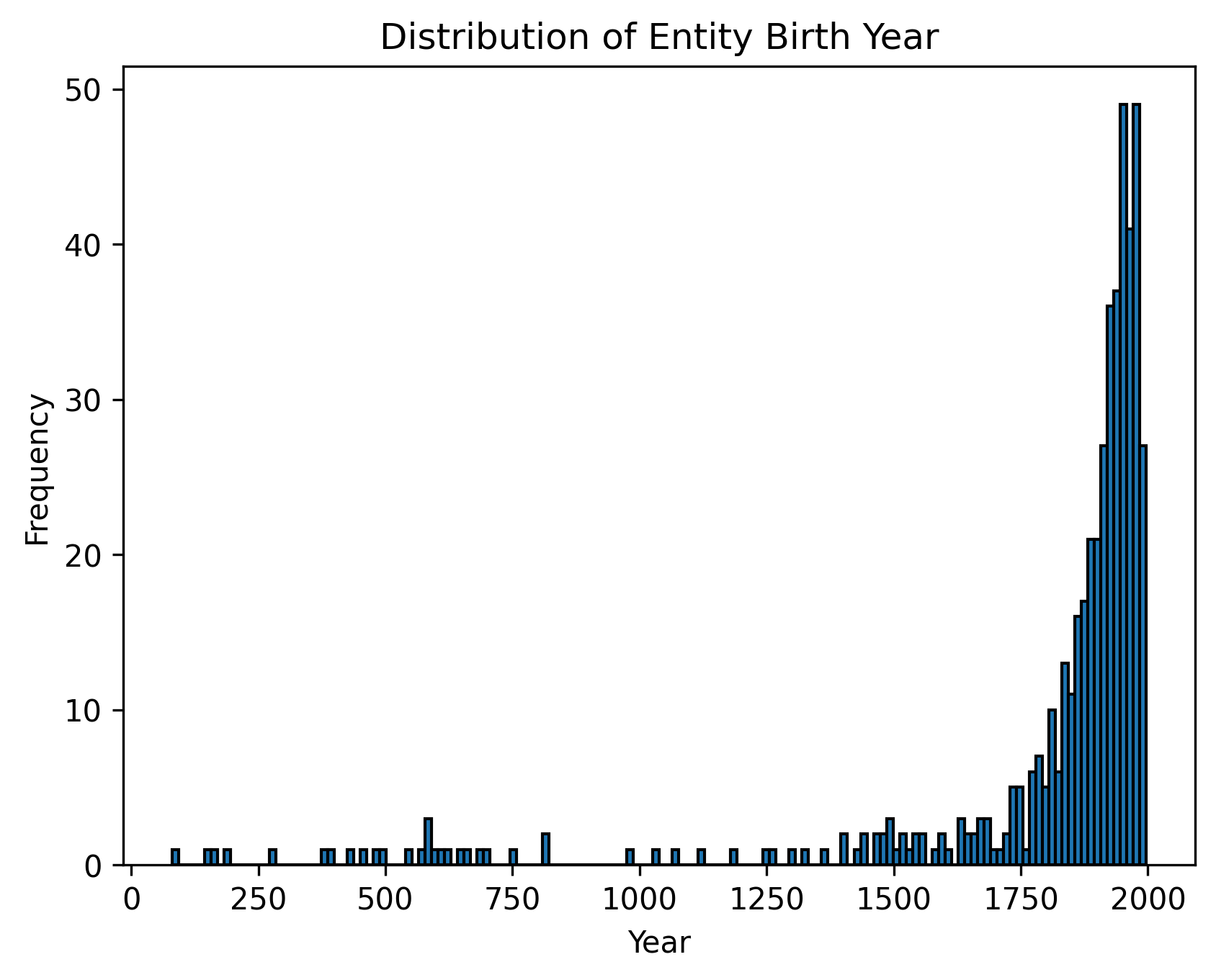}
        \caption{Birth Year}
        \label{fig:birth}
    \end{subfigure}

    \caption{Overview of entity diversity in the dataset. (a) Distribution of professions, (b) Geographic diversity of entities, (c) Birth year distribution}
    \label{fig:analysis}
\end{figure}
\subsection{Representation in the Data}
To assess the representation in our data, we analyzed the properties of the 500 original extracted entities. This analysis reveals the wide variety of professions included in the dataset, extracted using the Wikidata property $P106$, corresponding to the entity's profession. Fig. \ref{fig:profession} illustrates the diverse professions covered by the entities. The professions span a wide range, from actors, politicians, and writers to scientists, educators, and musicians. This wide range of professions ensures that the dataset captures a diverse spectrum of careers, making it more versatile and useful across various fields.

We also performed a country of citizenship ($P27$) analysis for the entities. Given that this is a multilingual dataset, it was important to ensure the inclusion of individuals from various backgrounds. Fig. \ref{fig:country} provides a breakdown of the countries represented. As shown, the dataset includes a diverse set of individuals from a range of regions, with representation from countries such as the United States, the United Kingdom, India, and several European countries. One thing to note is that the dataset includes representations from historical countries and regions that are now known by different names, due to the presence of historical entities within the data. This analysis highlights the wide geographical and cultural scope of the dataset, making it valuable in multilingual contexts, where some regions may not be as well represented. 

Finally, we also performed a birth year analysis for the extracted entities. This analysis provides valuable information on the temporal distribution of individuals in the dataset. Fig. \ref{fig:birth} illustrates the distribution of birth years, showing a clear concentration of individuals born in the nineteenth and twentieth centuries. As expected, this distribution suggests that the dataset is skewed toward modern figures, reflecting the increasing number of individuals whose information is accessible in current digital databases. However, there is still some representation from much earlier years, maintaining a sense of variety in the dataset and capturing a broader historical range. This balance between modern and earlier figures enhances the dataset's versatility while maintaining its relevance to current contexts.

\subsection{Final Dataset Distribution \& Analysis}
\begin{figure*}[htbp]
    \centering
    \begin{subfigure}[b]{0.32\textwidth}
        \centering
        \includegraphics[width=\textwidth]{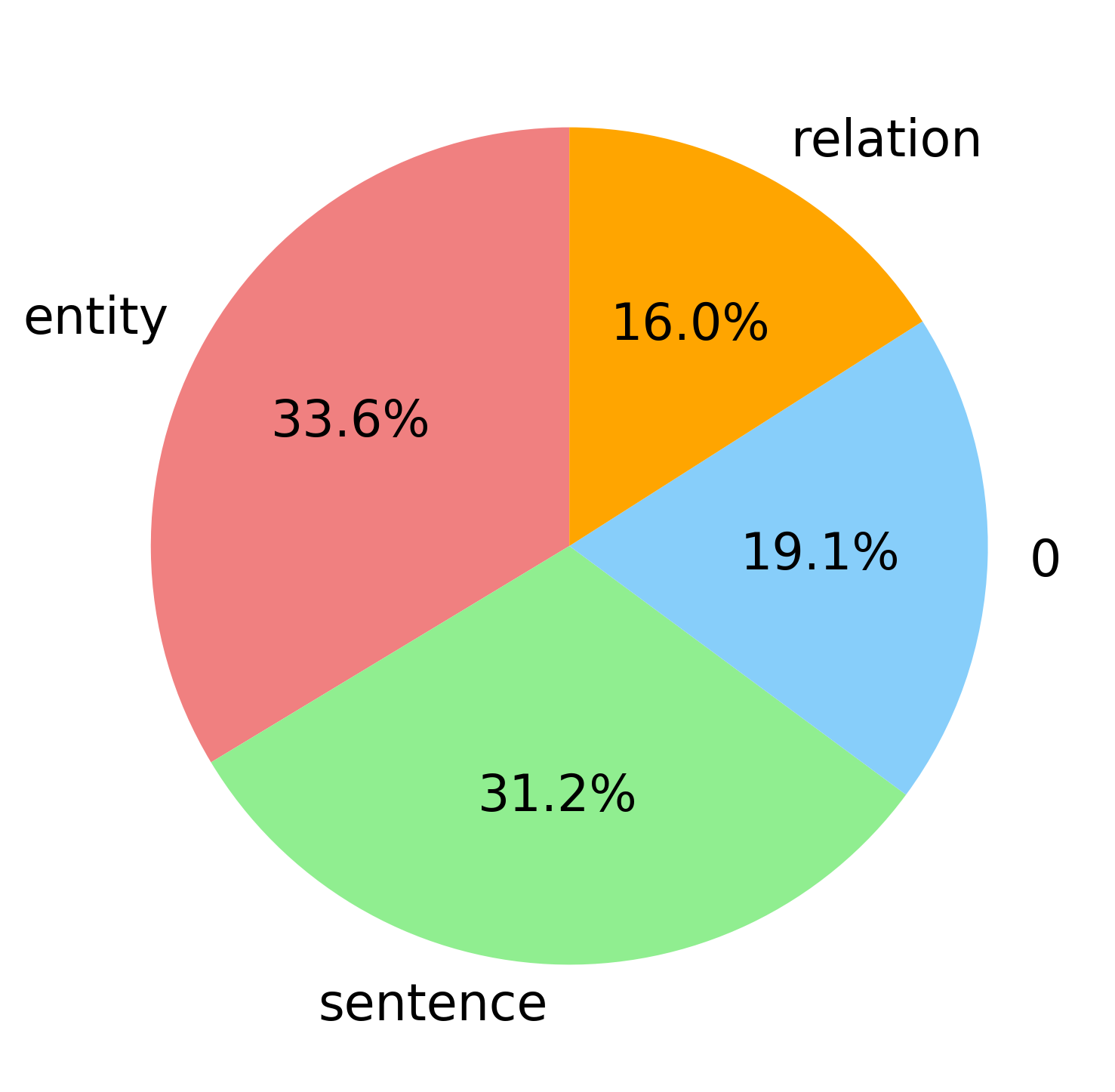}
        \caption{}
        \label{fig:en_dist}
    \end{subfigure}
    \hfill
    \begin{subfigure}[b]{0.32\textwidth}
        \centering
        \includegraphics[width=\textwidth]{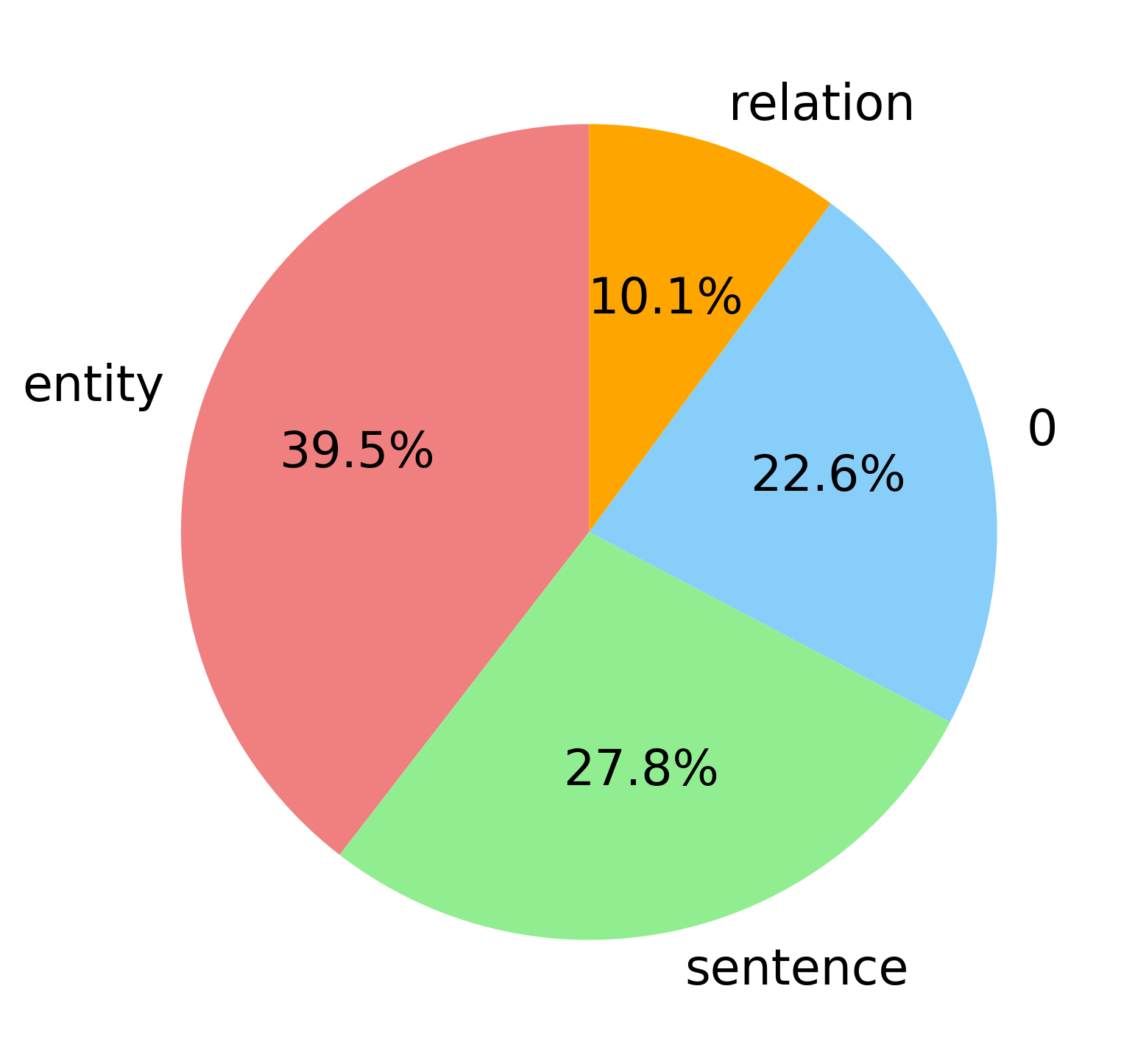}
        \caption{}
        \label{fig:ar_dist}
    \end{subfigure}
    \hfill
    \begin{subfigure}[b]{0.32\textwidth}
        \centering
        \includegraphics[width=\textwidth]{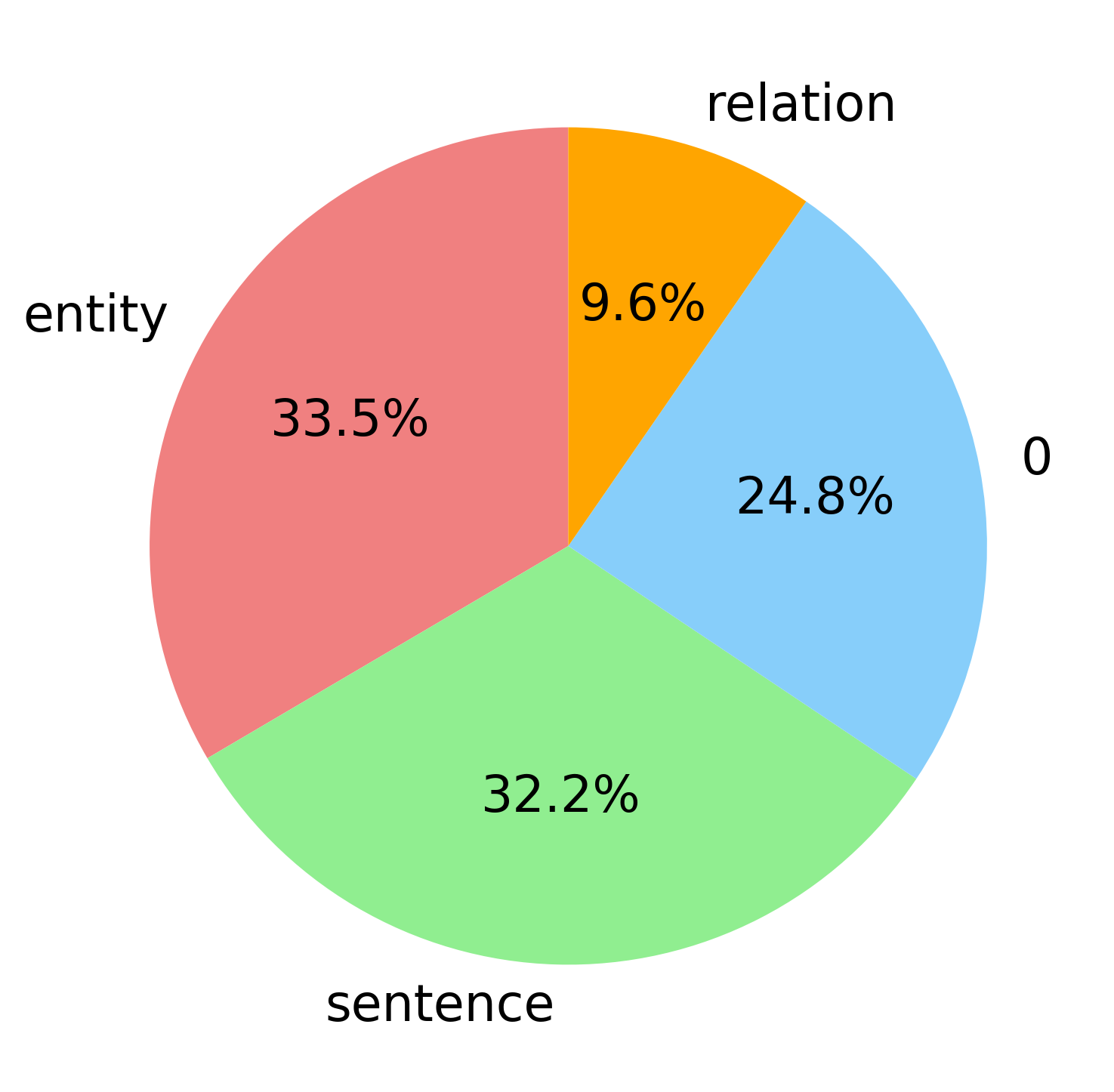}
        \caption{}
        \label{fig:tr_dist}
    \end{subfigure}
    \caption{Final dataset hallucination type distribution for (a) English, (b) Arabic, and (c) Turkish data}
    \label{fig:label_dist}
\end{figure*}



\begin{table}[]
\caption{Resulting dataset statistics for each language - comparing statistics of original sentences with the edited hallucinated sentences}
\resizebox{\columnwidth}{!}{%
\begin{tabular}{cccccc}
\toprule
\multirow{2}{*}{\textbf{Lang.}} & \multirow{2}{*}{\textbf{\# Sentences}} & \multicolumn{2}{c}{\textbf{\# Tokens}} & \multicolumn{2}{c}{\textbf{Av. Tokens/Sent.}} \\ 
                                &                                        & \textbf{Orig.}    & \textbf{Edited}    & \textbf{Orig.}        & \textbf{Edited}       \\ \midrule
\textbf{English}                & 1310                                   & 43,740            & 46,604             & 33.39                 & 35.58                 \\
\textbf{Arabic}                 & 828                                    & 24,482            & 25,416             & 29.57                 & 30.70                 \\
\textbf{Turkish}                & 978                                    & 23,071            & 25,396             & 23.59                 & 25.97      \\ \bottomrule          
\end{tabular} %
}
\label{tab:dataset_stats}
\end{table}
The distribution of hallucination types, as annotated by human annotators for each language, is illustrated in Fig. \ref{fig:label_dist}. These pie charts show the distribution of different types of hallucinations for the English, Arabic, and Turkish datasets. As seen, hallucinations are categorized into three primary types: entity, relation, and sentence. As discussed previously, some instances were labeled as \textbf{0}, indicating that there was no hallucination or if there were any grammatical errors.

For clarity, instances where the label was \textbf{0} (i.e., no hallucination detected), we exclude them from the following analysis. This ensures that the focus remains on the sentences that were appropriately modified with the required hallucinations.

After filtering out these non-hallucinated instances, the resulting dataset statistics are shown in Table \ref{tab:dataset_stats}. The table provides the following key metrics for each language:
\begin{itemize}
    \item \textbf{Number of Sentences}: The total number of sentences for each language.
    \item \textbf{Number of Tokens}: The total number of tokens for both the original and edited sentences.
    \item \textbf{Average Tokens per Sentence}: The average number of tokens per sentence for both the original and edited sentences.
\end{itemize}

This analysis highlights the differences between the original and edited sentences, revealing how the hallucinations impact the overall structure and complexity of the text. The average number of tokens per sentence for edited sentences tends to be higher, reflecting the linguistic complexity potentially introduced by hallucinations.

\section{Experiments}
\label{sec:exp}

\subsection{Experimental Setup}
\paragraph{Evaluation Models.} We use several state-of-the-art LLMs for evaluation. These include open-source models phi-4 (14b) \cite{abdin2024phi4technicalreport}, Mistral-7b-instruct \cite{jiang2023mistral7b}, qwen 2.5 (7b and 72b) \cite{qwen2025qwen25technicalreport}, gemini-flash-1.5-8b \cite{geminiteam2024gemini15unlockingmultimodal}, gemma-2-27b \cite{Riviere2024Gemma2I}, and llama-3.3-70b \cite{grattafiori2024llama3herdmodels}. In addition, close-source models including PaLM 2 \cite{anil2023palm2technicalreport}, and OpenAI's gpt-4o-mini and gpt-4o \cite{openai2024gpt4technicalreport} are explored. These experiments were conducted through pure prompting of the LLMs with no fine-tuning or access to internal hidden states. 

\paragraph{Implementation Details.} 
All implementations, including the generation of hallucinated data through gpt-4 and the evaluation of the data, were conducted using openrouter\footnote{https://openrouter.ai/} API access. For generation, the temperature was set to 1.0 and maximum number of tokens was 256. In the evaluation process, since the intended output is one of three labels, we set the temperature to 0 to ensure that it is deterministic.

\subsection{Results}

\begin{table*}[ht]
\centering
\caption{Performance comparison of models across English, Arabic, and Turkish.}

\resizebox{\textwidth}{!}{%
\begin{tabular}{lcccccccccccc}
\toprule
\textbf{Models} & \multicolumn{4}{c}{\textbf{English}} & \multicolumn{4}{c}{\textbf{Arabic}} & \multicolumn{4}{c}{\textbf{Turkish}} \\
\cmidrule(lr){2-5} \cmidrule(lr){6-9} \cmidrule(lr){10-13}
                & \textbf{Ent} & \textbf{Rel} & \textbf{Sent} & \textbf{Overall} 
                & \textbf{Ent} & \textbf{Rel} & \textbf{Sent} & \textbf{Overall} 
                & \textbf{Ent} & \textbf{Rel} & \textbf{Sent} & \textbf{Overall} \\ 
\midrule
phi-4                    & 94.13 & 95.37 & 20.36 & 65.88 & 92.67 & 87.96 & 29.63 & 69.44 & 91.26 & 86.40 & 21.53 & 60.84 \\
mistral-7b-instruct-v0.3 & 82.02 & 79.54 & 0.79 & 50.15 & 54.37 & 90.74 & 1.01 & 39.98 & 82.76 & 80.80 & 0.48 & 47.34 \\
qwen-2.5-7b-instruct     & 68.81 & 88.42 & 13.83 & 51.45 & 59.81 & 100.00 & 6.06 & 45.77 & 50.80 & 93.60 & 11.48 & 39.47 \\
gemini-flash-1.5-8b      & 95.96 & 93.44 & 14.23 & 63.89 & 91.49 & 95.37 & 3.37 & 60.39 & 96.78 & 88.00 & 6.22 & 56.95 \\
gemma-2-27b-it           & 83.30 & 71.43 & 12.65 & 53.66 & 81.09 & 74.07 & 7.41 & 53.74 & 79.54 & 76.00 & 12.68 & 50.51 \\
qwen-2.5-72b-instruct    & 86.06 & 96.53 & 31.23 & 66.95 & 81.56 & 97.22 & 30.30 & 65.22 & 85.06 & 92.80 & 39.23 & 66.46 \\
llama-3.3-70b-instruct   & 98.53 & 93.05 & 12.65 & 64.27 & 97.40 & 84.26 & 13.80 & 65.70 & 92.87 & 85.60 & 11.72 & 57.26 \\
\hline
PaLM 2          & 93.94 & 95.37 & 2.37 & 58.85 & 89.60 & 94.44 & 3.37 & 59.30 & 94.02 & 96.0 & 2.63 & 55.21 \\
gpt-4o-mini     & 99.63 & 93.05 & 9.49 & 63.51 & 96.93 & 87.96 & 15.49 & 66.55 & 97.93 & 85.60 & 6.22 & 57.16 \\
gpt-4o             & 99.27 & 94.59 & 51.58 & 79.92 & 95.74 & 90.74 & 58.92 & 81.88 & 97.47 & 91.20 & 47.61 & 75.36 \\

\bottomrule
\end{tabular}%
}
\label{tab:performance_comparison}
\end{table*}

To evaluate the performance of LLMs in identifying different types of hallucinations across multilingual settings, we provided the LLMs with the original and edited pairs of sentences, in addition to three examples corresponding to each of the types of hallucination. The evaluation prompt used for this can be found in Table \ref{tab:experiment_prompt} in Appendix \ref{app:prompts}. 

We report the accuracy of the classifications in Table \ref{tab:performance_comparison}. GPT-4o leads in all languages and hallucination types, which is  expected given that the data was generated using GPT-4o. However, phi-4 also shows strong performance, particularly in entity and relation hallucinations. Phi-4 achieved an overall accuracy of 65.88\% in English, 69.44\% in Arabic, and 60.84\% in Turkish, showcasing its competitive capabilities. Phi-4 outperformed models such as mistral-7b-instruct-v0.3 and qwen-2.5-72b-instruct, which showed lower overall performance, particularly in sentence hallucinations. An interesting and relatively surprising observation made is the fact that performance across most of the LLMs increases when presented with Arabic data, compared to other languages. 

The results for sentence hallucinations reveal a consistent trend across all languages: models generally performed the worst in this category. This is to be expected, as sentence-level errors are often longer and more complex, spanning entire sentences. These types of hallucinations are also more subjective to identify, as they require a deeper understanding of the sentence's context, structure, and meaning. Consequently, all models struggled more with sentence hallucinations compared to entity or relation hallucinations, which are more isolated and easier to identify. This emphasizes the need for further research and model improvements in detecting sentence hallucinations, as this issue clearly had a negative impact on the overall performance of all models. 

\begin{figure*}[htbp]
    \centering
    \begin{subfigure}[b]{0.32\textwidth}
        \centering
        \includegraphics[width=\textwidth]{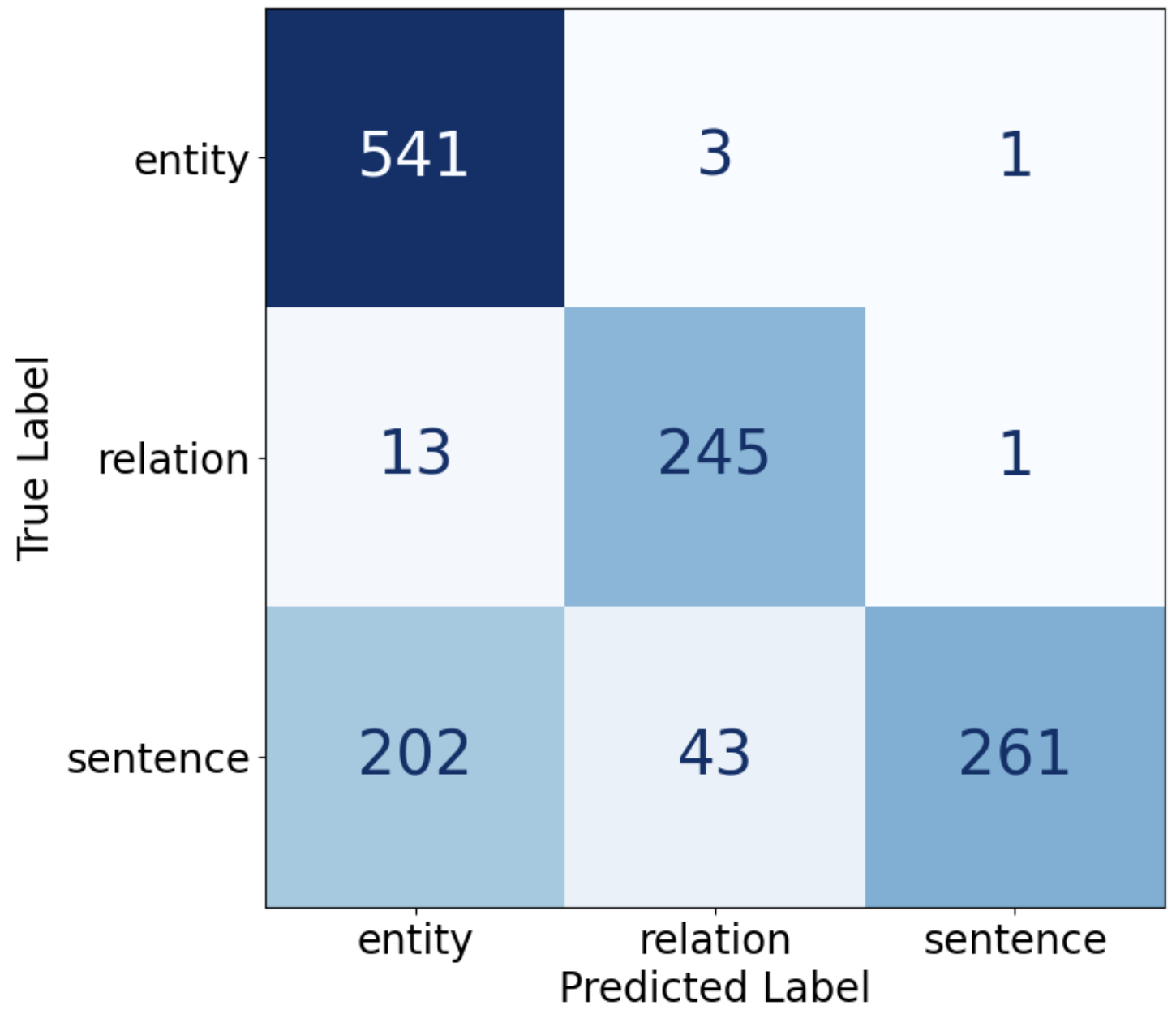}
        \caption{}
        \label{fig:sub1}
    \end{subfigure}
    \hfill
    \begin{subfigure}[b]{0.32\textwidth}
        \centering
        \includegraphics[width=\textwidth]{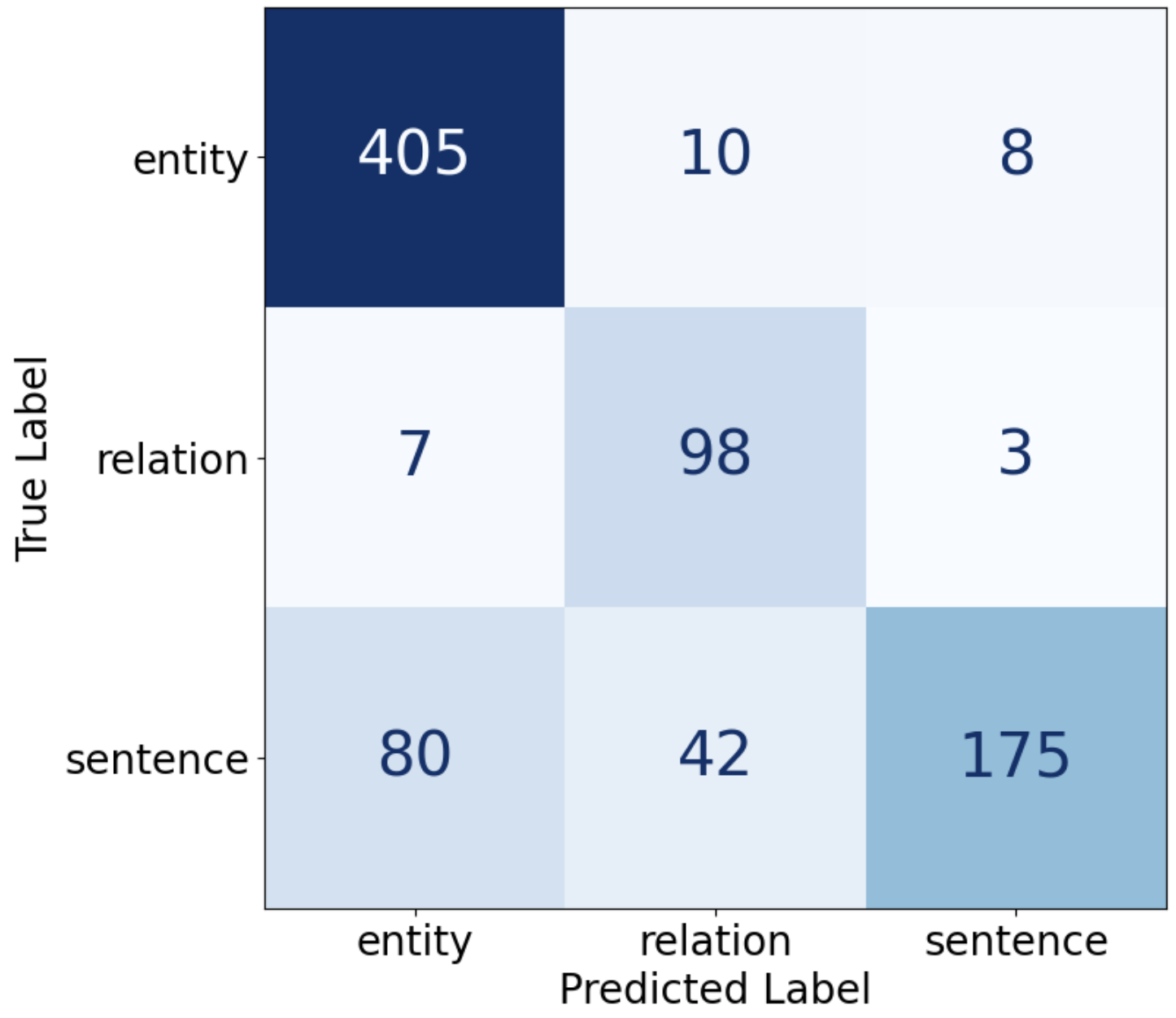}
        \caption{}
        \label{fig:sub2}
    \end{subfigure}
    \hfill
    \begin{subfigure}[b]{0.32\textwidth}
        \centering
        \includegraphics[width=\textwidth]{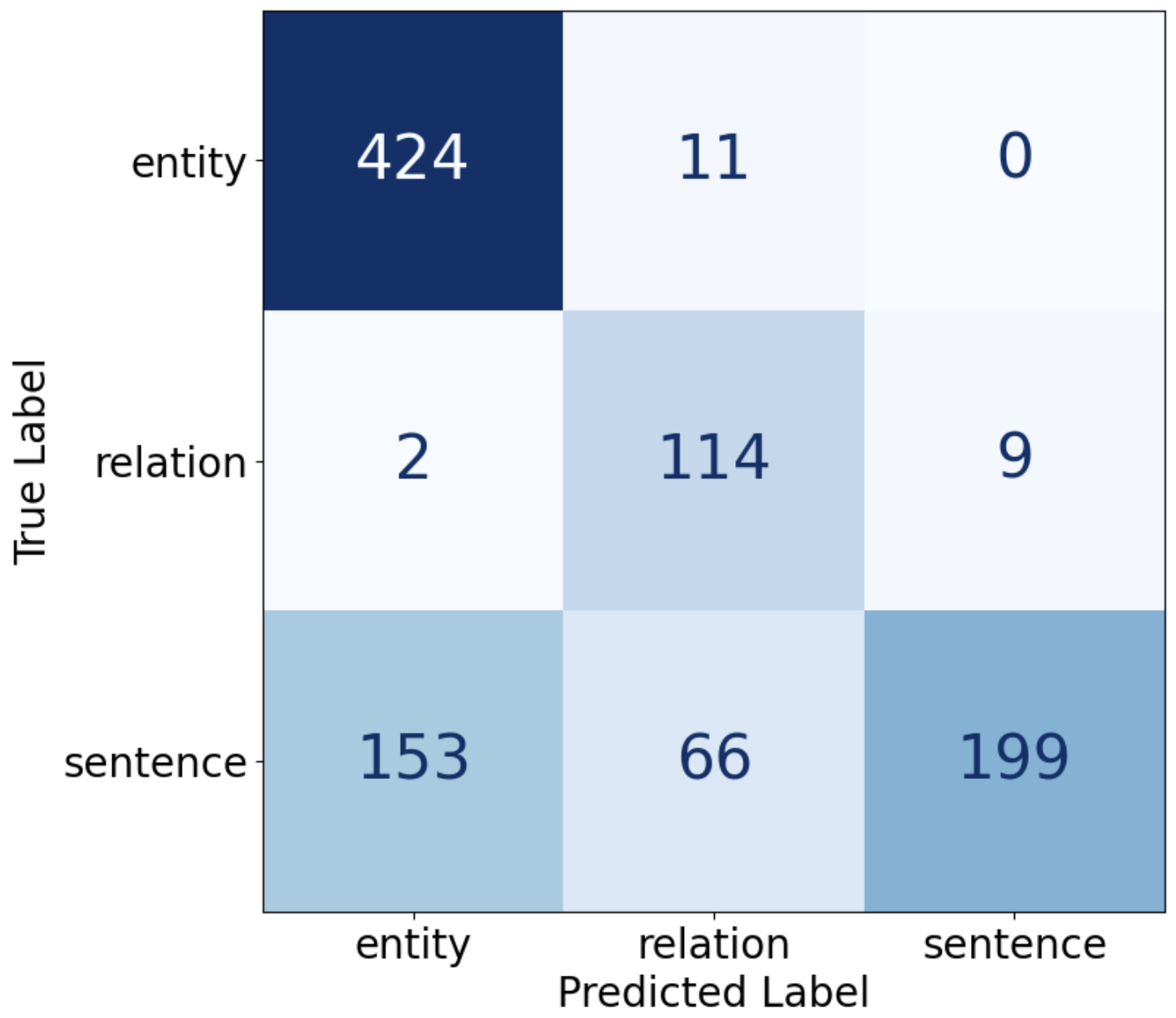}
        \caption{}
        \label{fig:sub3}
    \end{subfigure}
    \caption{Hallucination Category Confusion Matrices of gpt-4o labels for (a) English, (b) Arabic, and (c) Turkish data}
    \label{fig:confusion}
\end{figure*}

Digging deeper into GPT-4o's performance, we illustrate the confusion matrices for each language in Fig. \ref{fig:confusion}. Based on the classification predictions made by the LLM, as discussed previously, we see a clear performance drop in identifying sentence errors, as such errors seem to be misclassified as entity errors by the model, across all languages. 
To emphasize the uneven distribution of errors, we also report the number of samples in which the LLM fails to recognize the types of hallucinations in the languages explored in Table \ref{tab:failed_samples}. 
\begin{table}[]
\caption{Number of samples where GPT-4 fails to recognize each type of hallucination for all languages}

\begin{tabular}{lllll}
\toprule
\textbf{Language}   & \textbf{\# Failed} & \textbf{Ent.} & \textbf{Rel.} & \textbf{Sent.} \\ \midrule
\textbf{English} & 263                & 4             & 14            & 245            \\
\textbf{Arabic}  & 150                & 18            & 10            & 122            \\
\textbf{Turkish} & 241                & 11            & 11            & 219            \\ \bottomrule
\end{tabular}
\label{tab:failed_samples}
\end{table}

\section{Conclusion}
\label{sec:conclusion}
In this paper, we present \textit{HalluVerse25}, a fine-grained multilingual dataset for LLM-generated hallucinations, constructed by injecting hallucinated content using LLMs. To create this dataset, we collect biographical data from well-known figures from Wikipedia and extract factual sentences that are then used as input for an LLM to generate hallucinated text. \textit{HalluVerse25} is the first self-contained multilingual dataset that categorizes fine-grained hallucination types across multiple languages. To ensure high data quality, we implement a rigorous human annotation protocol for verification. Additionally, we conduct a detailed evaluation of LLM performance on the dataset, providing insights into how well these models identify different types of hallucinations in multilingual settings. We hope that \textit{HalluVerse25} will serve as a valuable resource for advancing research on LLM hallucinations in diverse linguistic contexts.

\section{Limitations}
The data set is collected on the basis of the availability of biographical information hosted on Wikidata and Wikipedia. Although we show in Section \ref{sec:analysis} that there is a decent amount of representation in terms of professions and countries, we must acknowledge the fact that it is unlikely that such platforms represent an accurate distribution of individuals in the world. For future work, our aim is to fill this gap by controlling the distribution in the data extraction process to clearly ensure a fair and distributed representation in the output. 

In addition, this dataset focuses on biographical data. Although this is a promising start in terms of creating multilingual benchmarks for fine-grained hallucination in LLMs, it is important to improve the versatility of the publicly available datasets by including other tasks as well, such as summarization, question answering, and more. Future work will build on this by exploring several multilingual tasks to add to this benchmark.

\newpage

\appendix
\section{SPARQL Query}
\label{app:sparql}

\begin{lstlisting}
SELECT DISTINCT ?item ?itemLabel ?enLink ?arLink ?trLink WHERE {
  ?item wdt:P31 wd:Q5.  \# Ensure the item is a human (Q5)
  
  \# Optional: Fetch Wikipedia links in various languages
  OPTIONAL { ?item wdt:P214 ?enLink. }  \# English Wikipedia link
  OPTIONAL { ?item wdt:P2025 ?arLink. }  \# Arabic Wikipedia link
  OPTIONAL { ?item wdt:P2025 ?trLink. }  \# Turkish Wikipedia link
  
  \# Ensure there are links in the specified languages
  FILTER(BOUND(?enLink) || BOUND(?arLink) || BOUND(?trLink))
  
  SERVICE wikibase:label { bd:serviceParam wikibase:language "[AUTO_LANGUAGE],en,ar,tr". }
}
LIMIT 500
\end{lstlisting}

\clearpage
\onecolumn
\section{Annotation Protocol}
\label{app:protocol}

\subsection*{Objective}
The objective of this task is to review pairs of sentences: an \textbf{original sentence} (ground truth) and an \textbf{edited sentence} (potentially containing an error). Annotators will determine if the edited sentence contains any factual inaccuracies or errors compared to the original sentence. If an error is present, annotators will classify the type of error as one of \textbf{[entity, relation, error]}. If the edited sentence is factually accurate (same as original sentence) OR the original sentence does not contain fact-checkable content to begin with OR the edited sentence contains grammatical errors, annotators should label it as \textbf{0}.

\subsection*{Annotation Steps}

\begin{enumerate}
    \item \textbf{Review the original and edited sentence pairs}:
    \begin{itemize}
        \item The \textbf{original sentence} contains the correct ground-truth information.
        \item The \textbf{edited sentence} may or may not contain an error.
    \end{itemize}
    
    \item \textbf{Determine if the edited sentence contains an error}:
    \begin{itemize}
        \item Compare the edited sentence to the original sentence to determine if an error has been introduced.
        \item If the edited sentence is factually accurate OR does not contain fact-checkable information, label it as \textbf{0}.
    \end{itemize}
    
    \item \textbf{If the edited sentence contains an error, classify the error type}:
    \begin{itemize}
        \item Use the appropriate error tags as described below, based on the types provided.
    \end{itemize}
\end{enumerate}

\subsection*{Error Types}
Annotators should use the following tags to classify errors in the edited sentence. For each type, three examples are provided to clarify what constitutes each error.

\begin{itemize}
    \item \textbf{0 (No Error or Non-Factual Statement)}: 
    \begin{itemize}
        \item If the edited sentence contains grammatical errors, non-factual content (e.g., opinions or subjective statements), or is similar to the original sentence, label it as \textbf{0}.
        \item \textbf{Example 1:}
            \begin{itemize}
                \item Original: “John Doe is an experienced leader.”
                \item Edited: “John Doe is highly regarded for his leadership.” 
            \end{itemize}
        \item \textbf{Example 2:}
            \begin{itemize}
                \item Original: “John Doe’s work has been influential.”
                \item Edited: “John Doe has positively impacted many.”
            \end{itemize}
        \item \textbf{Example 3:}
            \begin{itemize}
                \item Original: “John Doe is going to run for President”
                \item Edited: “John Doe is going to run with President.”
            \end{itemize}
    \end{itemize}

    \item \textbf{Entity Error}: Incorrect entities (e.g., people, places, dates, or objects) in the edited sentence. 
    \begin{itemize}
        \item \textbf{Example 1:}
            \begin{itemize}
                \item Original: “John Doe was born in 1990.”
                \item Edited: “John Doe was born in 1995.”
            \end{itemize}
        \item \textbf{Example 2:}
            \begin{itemize}
                \item Original: “John Doe worked at Microsoft.”
                \item Edited: “John Doe worked at Apple.”
            \end{itemize}
        \item \textbf{Example 3:}
            \begin{itemize}
                \item Original: “John Doe graduated from Harvard.”
                \item Edited: “John Doe graduated from Yale.”
            \end{itemize}
    \end{itemize}

    \item \textbf{Relation Error}: Incorrect semantic relationships between entities, actions, or prepositions in the edited sentence.
    \begin{itemize}
        \item \textbf{Example 1:}
            \begin{itemize}
                \item Original: “John Doe became the CEO of Google in 2015.”
                \item Edited: “John Doe joined Google as an engineer in 2015.”
            \end{itemize}
        \item \textbf{Example 2:}
            \begin{itemize}
                \item Original: “John Doe led the research team at IBM.”
                \item Edited: “John Doe worked under the research team at IBM.”
            \end{itemize}
        \item \textbf{Example 3:}
            \begin{itemize}
                \item Original: “John Doe collaborated with Dr. Smith on the project.”
                \item Edited: “John Doe supervised Dr. Smith on the project .”
            \end{itemize}
    \end{itemize}

    \item \textbf{Sentence Error}: The entire edited sentence is factually incorrect or contradicts established information.
    \begin{itemize}
        \item \textbf{Example 1:}
            \begin{itemize}
                \item Original: “John Doe published his first paper in 2010.”
                \item Edited: “John Doe invented the theory of relativity. ”
            \end{itemize}
        \item \textbf{Example 2:}
            \begin{itemize}
                \item Original: “John Doe received the Science Award in 2018.”
                \item Edited: “ John Doe won an Academy Award for acting in 2019. ”
            \end{itemize}
        \item \textbf{Example 3:}
            \begin{itemize}
                \item Original: “John Doe was recognized for his research in biology.”
                \item Edited: “John Doe was recognized for his achievements in film directing.”
            \end{itemize}
    \end{itemize}
\end{itemize}

\clearpage
\onecolumn
\section{Prompts used for \textit{HalluVerse25} Construction Pipeline}
\label{app:prompts}
\noindent\rule{\linewidth}{0.4mm} 

\begin{longtable}{|p{0.95\textwidth}|} 
    \hline
    \rowcolor{blue!10} 
    \textbf{Prompt:} Return a python list of 5 factual sentences from the following biography. Just return the list with no other text, and do not make any edits to any of the sentences: \\
    \hline
    \rowcolor{green!10} 
    \multicolumn{1}{|p{0.95\textwidth}|}{
    \textbf{Biography:} Ray Douglas Bradbury (August 22, 1920 – June 5, 2012) was an American author and screenwriter. One of the most celebrated 20th-century American writers, he worked in a variety of genres, including fantasy, science fiction, horror, mystery, and realistic fiction. \newline
    Bradbury is best known for his novel \textit{Fahrenheit 451} (1953) and his short-story collections \textit{The Martian Chronicles} (1950), \textit{The Illustrated Man} (1951), and \textit{The October Country} (1955). \newline
    Other notable works include the coming-of-age novel \textit{Dandelion Wine} (1957), the dark fantasy \textit{Something Wicked This Way Comes} (1962), and the fictionalized memoir \textit{Green Shadows, White Whale} (1992). He also wrote and consulted on screenplays and television scripts, including \textit{Moby Dick} and \textit{It Came from Outer Space}. Many of his works were adapted into television and film productions as well as comic books. Bradbury also wrote poetry, which has been published in several collections, such as \textit{They Have Not Seen the Stars} (2001). \newline
    Bradbury was born on August 22, 1920, in Waukegan, Illinois, to Esther (née Moberg) Bradbury (1888–1966), a Swedish immigrant, and Leonard Spaulding Bradbury (1890–1957), a power and telephone lineman of English ancestry. He was given the middle name "Douglas" after actor Douglas Fairbanks. \newline
    Bradbury was surrounded by an extended family during his early childhood and formative years in Waukegan. An aunt read him short stories when he was a child. This period provided foundations for both the author and his stories. In Bradbury's fiction, 1920s Waukegan becomes Green Town, Illinois.
    } \\
    \hline
    \caption{Instruction for Extracting Factual Sentences from a Biography}
    \label{tab:instruct_fact}
\end{longtable}

\clearpage
\noindent\rule{\linewidth}{0.4mm} 

\begin{longtable}{|p{0.95\textwidth}|} 
    \hline
    \rowcolor{blue!10} 
    \multicolumn{1}{|p{0.95\textwidth}|}{
    \textbf{Prompt:} Edit the sentence below by inserting an entity hallucination error in the sentence. \newline
    Replace the original sentence with the edited sentence, and clearly mark the edited sentence with the tag <entity\_error> </entity\_error>. DO NOT MODIFY anything else in the text: \newline
    \newline
    \textbf{Entity Errors:} Incorrect entities in statements that, if corrected, make the sentence factually accurate. Make sure you edit the sentence appropriately before tagging it. Some examples of this error include: \newline
    \newline
    \textbf{Example 1:} \newline
    Original: "Albert Einstein was born in Ulm, Germany, in 1879." \newline
    Edited: <entity\_error>Albert Einstein was born in Vienna, Austria, in 1879.</entity\_error> \newline
    \newline
    \textbf{Example 2:} \newline
    Original: "Marie Curie won two Nobel Prizes in Chemistry." \newline
    Edited: <entity\_error>Marie Curie won two Nobel Prizes in Physics.</entity\_error> \newline
    \newline
    \textbf{Example 3:} \newline
    Original: "Neil Armstrong was the first person to land on Mars in 1969." \newline
    Edited: <entity\_error>Neil Armstrong was the first person to land on the Moon in 2000.</entity\_error> \newline
    \newline
    Return only the full edited sentence and no other text.
    } \\
    \hline
    \rowcolor{green!10} 
    \textbf{Original:} \{sentence\} \\
    \rowcolor{green!10} 
    \textbf{Edited:} \\
    \hline
    \caption{Instruction for Inserting Entity Hallucination Errors in Sentences}
    \label{tab:entity_prompt}
\end{longtable}

\clearpage
\onecolumn

\noindent\rule{\linewidth}{0.4mm} 

\begin{longtable}{|p{0.95\textwidth}|} 
    \hline
    \rowcolor{blue!10} 
    \multicolumn{1}{|p{0.95\textwidth}|}{
    \textbf{Prompt:} Edit the sentence below by inserting a relation hallucination error in the sentence. \newline
    Replace the original sentence with the edited sentence, and Clearly mark the edited sentence with the tag <relation\_error> </relation\_error>. DO NOT MODIFY anything else in the text: \newline
    \newline
    \textbf{Relation Errors:} Incorrect semantic relationships (e.g., verbs, prepositions) within the statement. Make sure you edit the sentence appropriately before tagging it. Some examples of this error include: \newline
    \newline
    \textbf{Example 1:} \newline
    Original: "He was awarded the Nobel Prize for his groundbreaking work in physics." \newline
    Edited: <relation\_error>He was awarded the Nobel Prize by his groundbreaking work in physics.</relation\_error> \newline
    \newline
    \textbf{Example 2:} \newline
    Original: "The team won the championship because they trained hard all season." \newline
    Edited: <relation\_error>The team won the championship so they trained hard all season.</relation\_error> \newline
    \newline
    \textbf{Example 3:} \newline
    Original: "The artist painted the portrait to capture the emotions of the subject." \newline
    Edited: <relation\_error>The artist painted the portrait to hide the emotions of the subject.</relation\_error> \newline
    \newline
    Return only the full edited sentence and no other text.
    } \\
    \hline
    \rowcolor{green!10} 
    \textbf{Original:} \{sentence\} \\
    \rowcolor{green!10} 
    \textbf{Edited:} \\
    \hline
    \caption{Instruction for Inserting Relation Hallucination Errors in Sentences}
    \label{tab:relation_prompt}
\end{longtable}

\clearpage

\begin{longtable}{|p{0.95\textwidth}|} 
    \hline
    \rowcolor{blue!10} 
    \multicolumn{1}{|p{0.95\textwidth}|}{
    \textbf{Prompt:} Edit the sentence below by inserting a sentence hallucination error in the sentence. \newline
    Replace the original sentence with the edited sentence, and Clearly mark the edited sentence with the tag <sentence\_error> </sentence\_error>. DO NOT MODIFY anything else in the text: \newline
    \newline
    \textbf{Sentence Errors:} Entire statements that contradict evidence. Make sure you edit the sentence appropriately before tagging it. Some examples of this error include: \newline
    \newline
    \textbf{Example 1:} \newline
    Original: "Isaac Newton developed the laws of motion and gravity in 1687." \newline
    Edited: <sentence\_error>Isaac Newton was the first person to walk on the moon in 1687.</sentence\_error> \newline
    \newline
    \textbf{Example 2:} \newline
    Original: "Marie Curie won Nobel Prizes in both Physics and Chemistry." \newline
    Edited: <sentence\_error>Marie Curie discovered the planet Pluto and won a Nobel Prize for it.</sentence\_error> \newline
    \newline
    \textbf{Example 3:} \newline
    Original: "The Wright brothers successfully flew the first airplane in 1903." \newline
    Edited: <sentence\_error>The Wright brothers invented the steam engine and successfully flew it in 1903.</sentence\_error> \newline
    \newline
    Return only the full edited sentence and no other text.
    } \\
    \hline
    \rowcolor{green!10} 
    \textbf{Original:} \{sentence\} \\
    \rowcolor{green!10} 
    \textbf{Edited:} \\
    \hline
    \caption{Instruction for Inserting Sentence Hallucination Errors in Sentences}
    \label{tab:sentence_prompt}
\end{longtable}

\noindent\rule{\linewidth}{0.4mm} 

\begin{longtable}{|p{0.95\textwidth}|} 
    \hline
    \rowcolor{blue!10} 
    \multicolumn{1}{|p{0.95\textwidth}|}{
    \textbf{Prompt:} Given the definitions below, classify the type of error in the edited sentence compared to the original. 
    \newline
    \newline
    \textbf{Definitions:} 
\newline
    - \textbf{Entity Errors:} The edited sentence contains incorrect entities that, if corrected, make the sentence factually accurate. 
\newline
\newline
    - \textbf{Relation Errors:} The edited sentence contains incorrect semantic relationships, such as verbs or prepositions, that alter the meaning. 
\newline
\newline
    - \textbf{Sentence Errors:} The entire edited sentence contradicts the evidence provided.
    \newline
    \newline
    Which error type best describes the issue in the edited sentence? Choose from [entity, relation, sentence] and return only the class.
    } \\
    \hline
    \rowcolor{green!10}
    \textbf{Original:} \{original\} \\
    \rowcolor{green!10}
    \textbf{Edited:} \{edited\} \\
    \hline
    \caption{Instruction for Classifying Error Types in Edited Sentences}
    \label{tab:experiment_prompt}
\end{longtable}

\end{document}